\documentclass{article}

\usepackage[table,xcdraw]{xcolor}
\usepackage[preprint]{neurips_2025}

\usepackage[utf8]{inputenc} % allow utf-8 input
\usepackage[T1]{fontenc}    % use 8-bit T1 fonts
\usepackage{hyperref}       % hyperlinks
\usepackage{url}            % simple URL typesetting
\usepackage{booktabs}       % professional-quality tables
\usepackage{amsfonts}       % blackboard math symbols
\usepackage[most]{tcolorbox}
\newtcolorbox{mybox}[2][]{colback=gray!10,title=#2,#1}
\usepackage{xspace}

\usepackage{nicefrac}       % compact symbols for 1/2, etc.
\usepackage{microtype}      % microtypography
\usepackage{multirow}
\usepackage{pgfplots}
\pgfplotsset{compat=1.18}
\usepackage{amsmath}
\usepackage{amssymb}
\usepackage{mathtools}
\usepackage{amsthm}
\usepackage{enumitem}
\usepackage{booktabs}
\usepackage{colortbl}  
\usepackage{graphicx}   
\usepackage{amsfonts}
\definecolor{myMidBlue}{RGB}{195, 225, 250}
\definecolor{myLightBlueFrame}{RGB}{97, 162, 220} 
\definecolor{mySoftBlue}{RGB}{205, 230, 255} 
\definecolor{myVeryLightBlue}{RGB}{220, 235, 255}   
\definecolor{myExtremelyLightBlue}{RGB}{235, 245, 255}
\definecolor{exactProjectPink}{HTML}{FF4081}
\usepackage{bbm}
\definecolor{mygray}{gray}{0.9}

\title{Geometric Prior-Guided Federated Prompt Calibration}

\author{
  \textbf{Fei Luo}\textsuperscript{1},
  \quad \textbf{Ziwei Zhao}\textsuperscript{2}, 
  \quad \textbf{Mingxuan Wang}\textsuperscript{3},
  \quad \textbf{Duoyang Li}\textsuperscript{4},
  \quad \textbf{Zhe Qian}\textsuperscript{5},\\
  \quad \textbf{Jiayi Tuo}\textsuperscript{6},
  \quad \textbf{Chenyue Zhou}\textsuperscript{7}
  \quad \textbf{Yanbiao Ma}\textsuperscript{3}\thanks{Corresponding author. Email: \texttt{ybma1998@ruc.edu.cn}},
  \\
  \textsuperscript{1}Jishou University, 
  \textsuperscript{2}Technical University of Munich, 
  \textsuperscript{3}Renmin University of China,\\
  \textsuperscript{4}Northwestern Polytechnical University,
  \textsuperscript{5}South China Agricultural University,\\
  \textsuperscript{6}University of Science and Technology of China,\\
  \textsuperscript{7}Nanjing University of Aeronautics and Astronautics\\
}
\begin{document}

\maketitle

\vspace{-0.3cm}
\begin{abstract}
Federated Prompt Learning (FPL) offers a parameter-efficient solution for collaboratively training large models, but its performance is severely hindered by data heterogeneity, which causes locally trained prompts to become biased. Existing methods, focusing on aggregation or regularization, fail to address this root cause of local training bias. To this end, we propose Geometry-Guided Text Prompt Calibration (GGTPC), a novel framework that directly corrects this bias by providing clients with a global geometric prior. This prior, representing the shape of the global data distribution derived from the covariance matrix, is reconstructed on the server in a privacy-preserving manner. Clients then use a novel Geometry-Prior Calibration Layer (GPCL) to align their local feature distributions with this global prior during training. Extensive experiments show GGTPC's effectiveness. On the label-skewed CIFAR-100 dataset ($\beta$=0.1), it outperforms the state-of-the-art by 2.15\%. Under extreme skew ($\beta$=0.01), it improves upon the baseline by 9.17\%. Furthermore, as a plug-and-play module on the domain-skewed Office-Home dataset, it boosts FedAvg's performance by 4.60\%. These results demonstrate that GGTPC effectively mitigates data heterogeneity by correcting the fundamental local training bias, serving as a versatile module to enhance various FL algorithms.
\end{abstract}
\vspace{-0.2cm}

\section{Introduction}
Federated Learning (FL) has garnered significant attention from both academia and industry for its ability to collaboratively train shared global models across multiple clients while preserving local data privacy\cite{mcmahan2017communication,kairouz2021advances,yang2019federated,li2020federated2,li2021survey,yin2021comprehensive,bonawitz2019towards}. In recent years, visual foundation models like CLIP\cite{CLIP}, BLIP\cite{BLIP} and BLIP2\cite{BLIP2} have achieved breakthroughs across numerous visual tasks due to their robust generalization capabilities. Consequently, securely and efficiently integrating such models into federated learning frameworks has emerged as a critical research direction\cite{fedclip,guo2023promptfl}. However, directly fine-tuning these large scale models on clients incurs substantial computational and communication overhead. To address this, parameter-efficient alternatives like Federated Prompt Learning (FPL) have emerged \cite{guo2023promptfl,qiu2024federated,feng2023learning}. This approach significantly boosts training efficiency by communicating and aggregating only lightweight prompt vectors across clients, rapidly becoming a research hotspot in parameter-efficient federated learning. Recent surveys and domain deployments further illustrate FL's breadth across robust training, continual on-device learning, sequential recommendation, oncology data sharing, and scientific collaboration ecosystems\cite{chen2025advances,li2025unleashing,li2025systematic,qi2025oncology,qi2025science}.

Despite its promising prospects, the core challenge of federated prompt learning is the widespread data heterogeneity among clients\cite{zhao2018federated,li2020federated,li2022federated,ye2023heterogeneous,huang2024federated}. This stems from the intrinsic mechanism of prompt learning: the method aims to find optimal prompts that center text embeddings precisely around the global visual feature center. In a federated setting, however, each client's data distribution is often biased (e.g. containing only partial categories or exhibiting severe class imbalance), resulting in a significant discrepancy between local data and the ideal global distribution\cite{fedrs}. When clients train prompts based solely on their own biased local data, the resulting text prompts inevitably deviate from the true global visual center. This performance degradation, caused by aggregating locally biased updates, severely limits the practical application of federated prompt learning\cite{fedlc,yoon2021fedmix}.

To address the aforementioned issues, we propose a direct solution: guiding clients towards unbiased local calibration by providing them with prior knowledge of the global data distribution in a privacy-preserving manner. This avoids optimization within the limited scope of local data. The core idea is to leverage the geometric properties of the embedding distribution to efficiently quantify and transmit this global prior information. Specifically, we define the geometric shape of each category in the embedding space as its primary data distribution direction (the eigenvector of the covariance matrix) and the extent to which it spreads across directions (the corresponding eigenvalue). This geometric prior accurately captures the intrinsic structure of the global data and inherently protects privacy, as its transmission involves no raw data.

Based on the aforementioned core principles, we propose a novel framework called Geometry-Guided Text Prompt Calibration (GGTPC). This framework comprises two key steps. First, a privacy-preserving aggregation strategy is designed on the server side. This involves collecting partial data to enrich the client's local statistical information (mean vectors and covariance matrices) in order to reconstruct an approximation of the global covariance matrix for each category, from which global geometric priors are extracted. Secondly, an innovative GeometryPrior Calibration Layer (GPCL) is introduced on the client side. This module performs random transformations on local visual embeddings within the feature space under the guidance of global geometric priors to avoid complex sample generation, thereby better aligning their distributions with the ideal global distribution. This process optimizes text prompts towards unbiased global centers and integrates seamlessly into training workflows, enabling efficient end-to-end calibration.

The main contributions are summarized as follows:

\begin{itemize}
    \item \textbf{Novel Calibration Perspective:} We propose addressing data heterogeneity in FPL from the perspective of text prompt embedding calibration for the first time. By introducing global geometric priors to correct local training bias, we provide a new research direction for this field.
    \item \textbf{Efficient Calibration Module:} We design a sample-generation-free Geometric Prior Calibration Layer (GPCL), which, combined with an inverse frequency sampling strategy, achieves efficient end-to-end unbiased calibration while effectively mitigating local class imbalance.
    \item \textbf{Versatility and Compatibility:} We verify that the proposed GGTPC method can serve as a plug-and-play module, seamlessly integrating into various mainstream federated learning algorithms and consistently improving their performance across different data heterogeneity scenarios, demonstrating its broad applicability and practical value.
\end{itemize}

\vspace{-0.2cm}

\section{Related Work}

\subsection{Data Heterogeneity in Federated Learning}
\label{subsec: 2.1}

The objective of federated learning (FL) is to collectively construct a robust global model without necessitating the sharing of local data by clients. However, in practical applications, data across clients is often non-IID (non-Independent and Identically Distributed), a phenomenon known as data heterogeneity\cite{zhao2018federated,li2022federated,ye2023heterogeneous}. This fundamental challenge gives rise to discrepancies between local model updates and the global optimization objective, with the potential to impede the convergence speed and ultimate performance of the global model to a considerable degree\cite{li2020federated, kairouz2021advances}. 

To address this challenge, existing research primarily follows two technical approaches. The first category comprises client-side regularization methods, which introduce additional constraints during local training to reduce discrepancies between local and global models. For instance, FedProx\cite{li2020federated} incorporates a proximity term to limit the magnitude of local updates, while SCAFFOLD\cite{SCAFFOLD} introduces control variables to mitigate “client drift” in local training. The second category focuses on server-side aggregation optimization, aiming to design smarter aggregation schemes to replace the naive FedAvg\cite{mcmahan2017communication}. For instance, FedOPT\cite{reddi2020adaptive} applies principles from adaptive optimizers like Adam to the server-side model aggregation process, dynamically adjusting global model updates, while FedNova\cite{fednova} and FedDC\cite{gao2022feddc} correct objective inconsistency and decouple local drift for non-IID data. While these methods have achieved significant success in traditional federated learning models, they typically operate on the entire set of model parameters. When applied directly to vision foundation models with billions of parameters, they impose substantial computational and communication burdens, prompting researchers to explore more parameter-efficient federated learning paradigms.

Beyond these representative baselines, recent studies explore cross-model aggregation and adaptive mutation to further counter non-IID effects, including stochastic mutation, multi-model cross-aggregation, and layer-wise recombination across models\cite{hu2024fedmut,hu2024fedcross,hu2024aggregation,wang2024fedcda}. Prototype- or curriculum-based calibration offers another path by distilling or clustering global knowledge to correct local bias\cite{qi2023cross,qi2025cross,liu2023cross,qi2022clustering,wang2023dafkd,zhao2024data}, while personalized objectives strive to jointly improve global generalization and client-specific performance\cite{meng2024improving}.

Parallel efforts focus on federated out-of-distribution and domain-generalized learning. Geometric alignment and causal intervention ideas\cite{ma2025geometric,qi2025federated,qi2025global,qi2024attentive,zhang2025federated} are complemented by prototype-based domain adaptation and graph transfer learning tailored to skewed distributions\cite{fu2025beyond,fu2025federated,liao2025federated,mai2024fgtl}. These advances underscore the importance of explicitly aligning local representations to a reliable global structure, motivating our geometric-prior-guided calibration.

\subsection{Prompt Learning in Vision-Language Models}
\label{subsec: 2.2}

Prompt Learning has emerged as a novel Parameter-Efficient Fine-Tuning (PEFT) technique, offering a lightweight solution for adapting large scale vision-language models (VLMs) to downstream tasks\cite{zhou2022learning,zang2022unified,khattak2023maple}. The fundamental principle of this approach entails the freezing of the majority of the parameters of a pre-trained model, whilst task execution is guided exclusively by the optimization of a small number of learnable prompt vectors.

CoOp\cite{zhou2022learning} first proposed learning continuous text prompt vectors to better adapt to downstream visual classification tasks. To address CoOp's insufficient generalization on new categories, CoCoOp\cite{zhou2022conditional} further introduced a meta-network to dynamically generate instance-conditional prompts, enhancing the model's generalization capability. Subsequent research has extended and refined prompt learning from various perspectives. For instance, MaPLe\cite{khattak2023maple} explored joint prompting for visual and linguistic branches to achieve multimodal synergy. Other works like LASP\cite{bulat2023lasp} and KgCoOp\cite{yao2023visual} focused on incorporating external knowledge or structural information into prompts for more robust performance, while recent approaches such as DePT\cite{zhang2024dept} and interactive prompting\cite{li2023efficient} further improve generalization via decoupled or multimodal prompt designs. However, these methods were developed under centralized settings, fundamentally assuming access to a complete dataset representing the global distribution. Consequently, they failed to address the challenge of learning unbiased prompts when data is distributed and heterogeneous.

\subsection{Federated Prompt Learning}
\label{subsec: 2.3}

Federated Prompt Learning (FPL) aims to combine the parameter efficiency of prompt learning with the privacy protection advantages of federated learning. PromptFL\cite{guo2023promptfl} pioneered the introduction of prompt modules into federated systems, replacing traditional model fine-tuning with collaborative training using textual prompts. Subsequent work has largely focused on optimizing prompts across different modalities. In the realm of visual prompts, methods like FedVPT\cite{fedvpt}, FedPR\cite{feng2023learning} and FedCLIP\cite{fedclip} explored learning client-specific or shared prompts within federated frameworks. For textual prompts, FedTPG\cite{qiu2024federated} designed a unified cross-client prompt generation network to learn more generalizable global prompts. While these approaches have advanced federated prompt learning, they primarily focus on refining aggregation strategies or designing novel prompt modules, often overlooking the fundamental issue stemming from data heterogeneity: due to significant distribution shifts between local and global data, prompts learned locally by clients are inherently biased. Existing methods fail to directly address this local training bias arising from distribution shifts.

In stark contrast to all the aforementioned approaches, the GGTPC framework proposed in this paper tackles the problem from a fundamentally different angle. Rather than focusing solely on improving aggregation algorithms or regularization terms, we directly confront the inherent distributional bias within the local training process. The core contribution of this paper is the pioneering proposal of privately transmitting a global geometric prior of the data distribution to each client. This enables clients to locally simulate a quasi-global, unbiased distributional environment. Consequently, it allows for direct calibration of the learning objectives for text prompts, fundamentally aligning them with the true global visual center.

\section{Methodology}
\label{sec: method}

In this section, we will elaborate on the Geometry-Guided Text Prompt Calibration (GGTPC) framework proposed in this paper. First, we formally define the problem and clarify its fundamental motivation. Next, we introduce the core concept of “embedding distribution geometry” and detail its acquisition method under privacy constraints in federated learning. Finally, we describe how clients leverage this geometric prior to perform unbiased local calibration through the Geometric Prior Calibration Layer (GPCL) and inverse frequency sampling strategy designed in this paper, and extend the framework to more challenging multi-domain scenarios.

\subsection{Problem Formulation and Motivation}

In standard vision-language models (such as CLIP), the objective of prompt learning is to optimize a learnable text prompt vector $p_{c}$ for category $c$. The goal is to align the embedding obtained through the text encoder $g_{t}(\cdot)$ with the \text{global} center $\mu_{c}^{\text{global}}$ of the visual features generated by the visual encoder $g_{v}(\cdot)$ for that category. Formally, this objective can be expressed as maximizing the cosine similarity:

\begin{equation}
\max \frac{g_t(p_c) \cdot \mu_c^{\text{\text{global}}}}{\|g_t(p_c)\|\|\mu_c^{\text{\text{global}}}\|}
\end{equation}
which $\mu_c^{\text{\text{global}}} = \mathbb{E}_{(x,c) \sim \mathcal{D}_{\text{\text{global}}}} [g_v(I_c)]$

However, in the federated learning setting, client $k$ can only access its local dataset $D_k$ and can only compute a biased local visual feature center $\,u_c^k$ from it. Due to data heterogeneity, a significant distribution skew exists between the local data distribution $P_{k}(c)$ and the \text{global} distribution $P_{\text{global}}(c)$, leading to:

\begin{equation}
\mu_c^k = \mathbb{E}_{I \sim \mathcal{D}_c^k} [g_v(I)] \neq \mu_c^{\text{\text{global}}}
\end{equation}

This discrepancy between local and \text{global} centers forms the core dilemma of federated prompt learning. Optimizing the text prompt $p_c$ solely on local data erroneously guides its embedding $g_t(p_c)$ toward a biased local center $\mu_c^k$, rather than its true target $\mu_c^{\text{global}}$. Aggregating these locally biased prompts inevitably yields a suboptimal \text{global} prompt. To overcome this fundamental challenge, knowledge about the \text{global} data distribution must be introduced to the client to correct the bias from local training. We contend that providing only the \text{global} mean (first-order moment) is insufficient. To fully characterize the distribution's structure, second-order moment information (covariance) is crucial. Therefore, this paper proposes utilizing the “Geometric Shape” derived from the covariance matrix as an ideal, privacy-preserving prior knowledge to guide clients toward unbiased calibration.

\subsection{Definition and Acquisition of \text{global} Geometric Priors}

We emphasize the use of the concept of "Geometric Shapes" rather than merely "covariance" because it provides a more intuitive and profound perspective for understanding data heterogeneity. The distribution of visual features in the embedding space is typically not an isotropic sphere, but rather an anisotropic ellipsoid with distinct directionality in high-dimensional space. These principal axes (feature vectors) reveal variations in intra-class physical properties (e.g., changes in pose or lighting), while their lengths (eigenvalues) quantify this diversity. Therefore, an unbiased text prompt must not only align with the distribution's center (mean) but also comprehend its structural morphology. The geometric prior in this paper is specifically designed to convey this structural information, guiding the prompt to learn the intrinsic manifold of \text{global} features rather than merely their average positions.

\subsubsection{Definition of Embedding Distribution Geometry}

We formally define the p-dimensional embedding of the geometric shape (GS) of category $c$ in space as the set of eigenvectors and eigenvalues obtained from the eigenvalue decomposition of its \text{global} visual feature covariance matrix $\sum_{c}^{\text{global}}$.

Specifically, for all \text{global} visual embeddings $\{v_1, . . . , v_n\}$ of a given category $c$, its \text{global} covariance matrix is:

\begin{equation}
\Sigma^{\text{\text{global}}}_{c}
= \mathbb{E}\big[(v - \mu^{\text{\text{global}}}_{c})(v - \mu^{\text{\text{global}}}_{c})^{T}\big]
\end{equation}

By performing an eigenvalue decomposition on this positive definite matrix, we obtain $\sum_{c}^{\text{global}} = U\Lambda U^{T}$, where $U = [u_1, . . . , u_p]$ is a standard orthogonal matrix composed of eigenvectors, and $\Lambda = diag(\lambda_1, . . . , \lambda_p)$ is a diagonal matrix composed of the corresponding eigenvalues. Therefore, the geometric shape $GS_c$ of category c is defined as:

\begin{equation}
GS_c = \{(u_j, \lambda_j)\}_{j=1}^p
\end{equation}

This definition possesses a clear physical interpretation: the eigenvectors $\{u_j\}$ represent the principal axes of the data distribution, indicating the directions of maximum variance; the corresponding eigenvalues $\{\lambda_j\}$ quantify the variance or spread along these axes. Collectively, the geometric shapes $GS_c$ comprehensively describe the orientation, scale, and morphology of the intrinsic ellipsoid underlying the \text{global} data distribution.

\subsubsection{Privacy-Preserving Acquisition of \text{global} Geometry}

In the federated learning paradigm, aggregating raw data from all clients to compute $\Sigma_{c}^{\text{global}}$ is strictly prohibited. Therefore, we designed a distributed method to accurately reconstruct the \text{global} covariance matrix from local statistics without compromising data privacy. The \text{global} covariance can be decomposed as the sum of intra-client covariance and inter-client covariance. Let $n_c^k$ denote the number of samples for category $c$ on client $k$, and $N_c = \sum_{k}n_c^k$ denote the total number of samples. The \text{global} mean is defined as $\mu_c^{\text{global}} = \frac{1}{N}\sum_{k=1}^{K}n_c^k\mu_c^k$. The \text{global} covariance matrix can then be reconstructed as:

\begin{equation}
\Sigma_{c}^{\text{\text{global}}}
= \frac{1}{N_c} \left(
    \sum_{k=1}^{K} n_c^{k} \Sigma_c^{k}
    + \sum_{k=1}^{K} n_c^{k}
    (\mu_c^{k} - \mu_c^{\text{\text{global}}})
    (\mu_c^{k} - \mu_c^{\text{\text{global}}})^{T}
\right)
\end{equation}

where $\mu_c^k$ and $\Sigma_c^k$ represent the local mean and local covariance matrix for client k, respectively.

To ensure the robustness of the estimation and mitigate noise from data-sparse clients, this paper employs a client selection strategy. For each category $c$, a subset of clients with the most samples for that category is selected, such that their cumulative sample count reaches 80\% of the total. Only these selected clients contribute to the global geometric computation for category $c$.

The acquisition process follows these steps:
\begin{enumerate}
    \item \textbf{Server Request:} The server requests local statistics for each category from all clients.
    
    \item \textbf{Client Computation and Upload:} Each client $k$ computes the triplet 
    $(n_c^{k}, \mu_c^{k}, \Sigma_c^{k})$ for every category $c$ present in its local data
    and securely uploads it to the server.
    
    \item \textbf{Server-side Reconstruction and Distribution:} The server aggregates
    statistics from the selected clients according to the selection policy, and computes
    the global mean $\mu_c^{\text{global}}$ and global covariance
    $\Sigma_c^{\text{global}}$ using formula~(5). It then performs a singular value
    decomposition on $\Sigma_c^{\text{global}}$ to extract the geometric prior
    $GS_c$ and distributes it to all participating clients in the current training round.
\end{enumerate}

\subsection{Geometry-Guided Local Calibration}

\subsubsection{Geometric Prior Calibration Layer (GPCL)}

Upon receiving the global geometric prior $GS_c$, the client can proceed with local calibration. This paper introduces the Geometric Prior Calibration Layer (GPCL), functioning as a feature space data augmentation module. Its core idea involves applying a random perturbation to each local visual embedding, sampled from a zero-mean distribution whose shape is defined by the global geometric prior GSc. This effectively simulates virtual samples around each local data point that conform to the global distribution's geometric morphology.

For an arbitrary visual embedding $X_c^k$ of class $c$ from client $k$, its calibrated version $X_c^{k'}$ is computed as follows:

\begin{equation}
X_c^{k'} = X_c^k + \sum_{m=1}^{p} \epsilon_m \sqrt{\lambda_m} u_m, \quad \text{where } \epsilon_m \sim \mathcal{N}(0, 1)
\end{equation}

The perturbation term $d = \sum_{m=1}^{p} \epsilon_m \sqrt{\lambda_m} u_m$ is a random vector sampled from the multivariate Gaussian distribution $\mathcal{N}(0, \Sigma_c^{\text{\text{global}}})$. This transformation enriches the local embeddings, expanding their distribution to cover the principal regions of the \text{global} feature space while preserving their original local information. This process guides the optimization of text prompts away from the biased local centers and toward a target that is more representative of the \text{global} distribution, thereby achieving unbiased calibration. The entire process functions as a lightweight pre-processing transformation, seamlessly integrated for end-to-end training.

\subsubsection{Mitigating Class Imbalance via Inverse Frequency Sampling}

Local class imbalance represents another critical issue, wherein majority classes dominate gradient updates. To address this, we employ an inverse frequency sampling strategy. For a client $k$, we first identify the maximum sample count $n_{\max}^k$ across all classes. Then, we define a sampling weight $w_c^k = n_{\max}^k / n_c^k$ for each class $c$, where $n_c^k$ is the number of samples in that class. Finally, the probability of sampling class $c$ is given by:

\begin{equation}
P_c^k = \frac{w_c^k}{\sum_{j=1}^C w_j^k}
\end{equation}

This strategy assigns a higher selection probability to classes with fewer samples, ensuring a more balanced representation of all classes in training batches and promoting fairer learning.

\subsection{Extension to Multi-Domain Scenarios}

In the more challenging multi-domain federated setting, where each client's data originates from a distinct domain, clients face a dual information gap: they lack not only knowledge of the \text{global} shape but also positional information regarding data from other domains. Even with a shared geometric shape, clients do not know where to "place" this shape in the embedding space to simulate data from unseen domains. To address this, we extend the GGTPC framework by introducing class prototypes as positional priors.

\begin{enumerate}
    \item \textbf{Distribution of Shape and Position Priors:} The server computes and distributes a shared \text{global} geometric prior $GS_c$ (assuming intra-class geometric similarity across domains). Additionally, it computes the mean embedding for each class $c$ in each domain $d$, referred to as the class prototype $\mu_c^d$. The server then sends the set of all out-of-domain prototypes $\{\mu_c^{d'} | d' \neq d_k\}$ to client $k$.
    
    \item \textbf{Collaborative Calibration:} The local training data pool of client $k$ is augmented with these received prototypes. During training, if a local sample $X_c^k$ is selected, it is calibrated using Equation (6). If an out-of-domain prototype $\mu_c^{d'}$ is sampled, it is similarly transformed via the GPCL mechanism: $\mu_c^{d''} = \mu_c^{d'} + \sum \epsilon_m \sqrt{\lambda_m} u_m$. This generates a virtual feature centered in the foreign domain but exhibiting the \text{global} shape.
    
    \item \textbf{Updating Sampling Strategy:} The inverse frequency sampling strategy is adjusted to include these prototypes, treating each prototype as a single sample with a correspondingly high weight, ensuring that both real and virtual data are sampled fairly for training.
\end{enumerate}

By combining positional and geometric priors, clients can effectively simulate a more complete \text{global} distribution spanning all domains, thereby enabling the learning of robust text prompts adapted to multi-domain environments.

\subsection{Overall Algorithm Flow}

The algorithm begins with the initialization of \text{global} parameters by the server. In each communication round, the server first collects local statistics (sample counts, means, and covariance matrices) from a selected subset of clients to calculate the \text{global} geometry. After aggregating this information to reconstruct the \text{global} covariance matrix, the server extracts the geometric prior via eigen decomposition and distributes it to all clients. Upon receiving the geometric prior, clients utilize the proposed Geometric Prior Calibration Layer (GPCL) and the inverse frequency sampling strategy during their local training process to calibrate their data and optimize, resulting in updated local prompts. Finally, the server aggregates the updates uploaded by all clients to generate a new \text{global} prompt and proceeds to the next communication round.

\section{Experiments}
\label{sec: exp}

This section presents a comprehensive series of experiments to systematically evaluate the effectiveness of the proposed GGTPC framework. Our evaluation is designed around three primary objectives: 1) To validate the performance of GGTPC in diverse and challenging data heterogeneity scenarios, including label skew, domain skew, and mixed skew where both coexist; 2) To demonstrate the capability of GGTPC as a plug-and-play module that can be seamlessly integrated into and consistently enhance various existing Federated Learning algorithms; 3) To provide intuitive visual evidence that, beyond improving accuracy metrics, GGTPC indeed achieves its goal of unbiased calibration of text prompt embeddings.

\begin{table*}[t]
\centering
\caption{Performance comparison on CIFAR-100 and Tiny-ImageNet datasets under different degrees of label skew ($\beta$). The best results are marked in \underline{\textbf{bold and underline}}. FedAvg (CoOp) denotes using CoOp as the backbone with the FedAvg algorithm.}
\vspace{1mm}
\setlength{\tabcolsep}{6pt}
\renewcommand{\arraystretch}{1.05}
\begin{small}
\begin{tabular}{r ccc ccc}
\toprule
\multirow{2}{*}{Method} & \multicolumn{3}{c}{CIFAR-100} & \multicolumn{3}{c}{Tiny-ImageNet} \\ 
\cmidrule(lr){2-4} \cmidrule(lr){5-7}
& 0.5 & 0.3 & 0.1 & 0.5 & 0.3 & 0.1 \\
\midrule
Zero-shot CLIP & \multicolumn{3}{c}{64.87} & \multicolumn{3}{c}{63.67} \\ 
\hline
FedVPT   & 83.53 & 83.18 & 80.99 & 75.91 & 75.67 & 74.30 \\
PromptFL & 75.05 & 75.36 & 72.19 & 72.66 & 71.51 & 67.85 \\
FedTPG   & 71.40 & 70.95 & 68.63 & 67.63 & 66.72 & 64.71 \\
FedPR    & 81.17 & 80.44 & 78.91 & 72.59 & 72.22 & 70.80 \\
FedCLIP  & 72.03 & 71.20 & 70.64 & 70.41 & 70.37 & 69.50 \\
\midrule
FedAvg (CoOp) & 79.28 & 77.72 & 75.92 & 76.84 & 75.42 & 73.69 \\
+ \textbf{GGTPC} & \underline{\textbf{84.21}} & \underline{\textbf{83.75}} & \underline{\textbf{83.14}} & \underline{\textbf{80.24}} & \underline{\textbf{79.93}} & \underline{\textbf{79.26}} \\
\bottomrule
\end{tabular}
\end{small}
\label{tab:1}
\vspace{-3mm}
\end{table*}

\subsection{Experimental Setup}

\textbf{Label-Skewed Datasets:} We evaluate GGTPC on three standard single-domain image classification benchmarks: CIFAR-10, CIFAR-100 \cite{krizhevsky2009learning}, and Tiny ImageNet-200 \cite{deng2009imagenet}.
\begin{itemize}
    \item CIFAR-10 and CIFAR-100 contain $32 \times 32$ images spanning 10 and 100 classes, respectively.
    \item Tiny ImageNet-200 is a subset of ImageNet containing 200 classes with an image size of $64 \times 64$.
\end{itemize}

Following recent benchmark settings, we simulate a federated environment comprising 10 clients. To introduce label skew, we partition the data among clients using a Dirichlet distribution $\text{Dir}(\beta)$\cite{fedavgm}. The concentration parameter $\beta > 0$ controls the degree of heterogeneity; a smaller $\beta$ value induces more severe class imbalance on each client, thereby creating a larger discrepancy between local and global distributions.

\begin{table*}[t]
\centering
\caption{Performance evaluation of GGTPC on CIFAR-10, CIFAR-100, and Tiny-ImageNet under severe label skew settings (i.e., smaller $\beta$ values).}
\vspace{1mm}
\setlength{\tabcolsep}{6pt}
\renewcommand{\arraystretch}{1.05}
\begin{small}
\begin{tabular}{r|ccccc}
\toprule
& \multicolumn{5}{c}{CIFAR-10} \\
\cmidrule(lr){2-6}
Method & 0.09 & 0.07 & 0.05 & 0.03 & 0.01 \\
\midrule
FedAvg (CoOp) & 93.32 & 92.80 & 92.21 & 91.69 & 90.91 \\
+ \textbf{GGTPC} & \underline{\textbf{96.62}} & \underline{\textbf{96.57}} & \underline{\textbf{96.11}} & \underline{\textbf{95.94}} & \underline{\textbf{95.66}} \\
\midrule
& \multicolumn{5}{c}{CIFAR-100} \\
\cmidrule(lr){2-6}
Method & 0.09 & 0.07 & 0.05 & 0.03 & 0.01 \\
\midrule
FedAvg (CoOp) & 75.87 & 74.15 & 72.99 & 70.86 & 69.71 \\
+ \textbf{GGTPC} & \underline{\textbf{80.74}} & \underline{\textbf{79.99}} & \underline{\textbf{79.63}} & \underline{\textbf{79.32}} & \underline{\textbf{78.88}} \\
\midrule
& \multicolumn{5}{c}{Tiny-ImageNet} \\
\cmidrule(lr){2-6}
Method & 0.09 & 0.07 & 0.05 & 0.03 & 0.01 \\
\midrule
FedAvg (CoOp) & 73.71 & 73.23 & 72.65 & 71.85 & 71.31 \\
+ \textbf{GGTPC} & \underline{\textbf{79.28}} & \underline{\textbf{78.81}} & \underline{\textbf{78.48}} & \underline{\textbf{78.15}} & \underline{\textbf{77.86}} \\
\bottomrule
\end{tabular}
\end{small}
\label{tab:2}
\vspace{-3mm}
\end{table*}

\begin{table*}[t]
\centering
\caption{Performance evaluation on the Digits dataset. AVG denotes the average accuracy, and STD denotes the standard deviation of accuracy.}
\vspace{1mm}
\setlength{\tabcolsep}{6pt}
\renewcommand{\arraystretch}{1.05}
\begin{small}
\begin{tabular}{r|cccc cc}
\toprule
Method & MNIST & SVHN & SYN & USPS & AVG $\uparrow$ & STD $\downarrow$ \\ 
\midrule
FedAvg & 90.40 & 34.68 & 46.99 & 60.30 & 58.09 & 20.74 \\
FedAvg (CoOp) & 83.05 & 66.79 & 86.27 & 80.62 & 79.18 & \underline{\textbf{7.43}} \\
+ \textbf{GGTPC} & 91.52 & 67.90 & 86.97 & 90.88 & \underline{\textbf{84.32}} & 9.64 \\
\midrule
SCAFFOLD & 97.79 & 26.64 & 90.69 & 94.45 & 77.39 & 29.41 \\
SCAFFOLD (CoOp) & 83.04 & 66.27 & 87.21 & 79.92 & 79.11 & \underline{\textbf{7.85}} \\
+ \textbf{GGTPC} & 91.32 & 67.67 & 86.97 & 90.43 & \underline{\textbf{84.10}} & 9.62 \\
\midrule
MOON & 92.78 & 33.36 & 39.28 & 68.11 & 58.36 & 23.82 \\
MOON (CoOp) & 89.69 & 66.23 & 86.82 & 88.24 & 82.74 & \underline{\textbf{9.59}} \\
+ \textbf{GGTPC} & 95.74 & 69.20 & 87.21 & 94.42 & \underline{\textbf{86.64}} & 10.58 \\
\midrule
FedDyn & 88.91 & 34.57 & 50.72 & 60.34 & 58.65 & 19.76 \\
FedDyn (CoOp) & 83.05 & 67.07 & 86.97 & 80.87 & 79.49 & \underline{\textbf{7.50}} \\
+ \textbf{GGTPC} & 92.43 & 67.60 & 86.92 & 91.58 & \underline{\textbf{84.63}} & 10.05 \\
\midrule
FedOPT & 92.71 & 31.32 & 87.92 & 87.62 & 74.89 & 25.38 \\
FedOPT (CoOp) & 81.17 & 66.94 & 87.31 & 81.02 & 79.11 & \underline{\textbf{7.47}} \\
+ \textbf{GGTPC} & 92.53 & 67.61 & 86.17 & 91.53 & \underline{\textbf{84.46}} & 10.02 \\
\midrule
FedProto & 90.54 & 34.61 & 58.00 & 89.54 & 68.18 & 23.38 \\
FedProto (CoOp) & 83.38 & 66.71 & 86.67 & 80.67 & 79.36 & \underline{\textbf{7.60}} \\
+ \textbf{GGTPC} & 91.66 & 67.63 & 86.72 & 91.28 & \underline{\textbf{84.32}} & 9.83 \\
\midrule
FedNTD & 52.31 & 18.03 & 97.29 & 58.07 & 56.43 & 28.12 \\
FedNTD (CoOp) & 86.42 & 65.88 & 85.92 & 81.46 & 79.92 & \underline{\textbf{8.33}} \\
+ \textbf{GGTPC} & 93.16 & 68.43 & 86.72 & 90.88 & \underline{\textbf{84.80}} & 9.73 \\
\bottomrule
\end{tabular}
\end{small}
\label{tab:3}
\vspace{-3mm}
\end{table*}

\textbf{Domain-Skewed Datasets:} We evaluate performance on two multi-domain datasets, Digits and Office-Home, where heterogeneity stems from domain shift.
\begin{itemize}
    \item \textbf{Digits} is a collection of four digit recognition datasets (MNIST \cite{lecun1998gradient}, SVHN \cite{netzer2011reading}, SYN \cite{ganin2015unsupervised}, USPS \cite{hull1994database}), each representing a unique domain.
    \item \textbf{Office-Home} \cite{venkateswara2017deep} contains images from four distinct domains: Art, Clipart, Product, and Real World, covering 65 shared object classes.
\end{itemize}
For these experiments, we adopt a strict "one-domain-one-client" assignment scheme, where each client is assigned data exclusively from one domain.

\begin{table*}[t]
\centering
\caption{Performance evaluation on the Office-Home dataset. A, C, P, and RW denote Art, Clipart, Product, and Real World domains, respectively.}
\vspace{1mm}
\setlength{\tabcolsep}{6pt}
\renewcommand{\arraystretch}{1.05}
\begin{small}
\begin{tabular}{r cccc cc}
\toprule
Method & A & C & P & RW & AVG $\uparrow$ & STD $\downarrow$ \\
\midrule
FedAvg (CoOp) & 82.65 & 73.90 & 91.99 & 91.48 & 85.01 & 7.41 \\
+ \textbf{GGTPC} & 89.28 & 80.84 & 94.76 & 93.58 & \underline{\textbf{89.61}} & \underline{\textbf{5.46}} \\
\midrule
SCAFFOLD (CoOp) & 82.71 & 74.59 & 92.12 & 92.01 & 85.36 & 7.30 \\
+ \textbf{GGTPC} & 89.45 & 81.72 & 95.11 & 93.92 & \underline{\textbf{90.05}} & \underline{\textbf{5.25}} \\
\midrule
MOON (CoOp) & 83.25 & 74.63 & 92.03 & 91.68 & 85.40 & 7.14 \\
+ \textbf{GGTPC} & 88.74 & 80.56 & 94.73 & 93.92 & \underline{\textbf{89.49}} & \underline{\textbf{5.64}} \\
\midrule
FedDyn (CoOp) & 83.80 & 76.88 & 92.42 & 91.21 & 86.08 & 6.25 \\
+ \textbf{GGTPC} & 89.55 & 82.06 & 95.47 & 93.81 & \underline{\textbf{90.22}} & \underline{\textbf{5.18}} \\
\midrule
FedOPT (CoOp) & 83.01 & 74.69 & 92.09 & 91.88 & 85.42 & 7.19 \\
+ \textbf{GGTPC} & 88.96 & 80.84 & 94.76 & 93.75 & \underline{\textbf{89.57}} & \underline{\textbf{5.50}} \\
\midrule
FedProto (CoOp) & 83.86 & 77.21 & 92.06 & 91.28 & 86.10 & 6.05 \\
+ \textbf{GGTPC} & 89.77 & 81.69 & 95.23 & 93.90 & \underline{\textbf{90.15}} & \underline{\textbf{5.28}} \\
\midrule
FedNTD (CoOp) & 84.46 & 77.14 & 92.26 & 91.68 & 86.38 & 6.16 \\
+ \textbf{GGTPC} & 89.50 & 82.26 & 95.38 & 94.30 & \underline{\textbf{90.36}} & \underline{\textbf{5.17}} \\
\bottomrule
\end{tabular}
\end{small}
\label{tab:4}
\vspace{-3mm}
\end{table*}

\begin{table*}[t]
\centering
\caption{Performance evaluation on the Office-Caltech-LDS dataset. Am, Ca, D, and W denote Amazon, Caltech, DSLR, and Webcam domains, respectively.}
\vspace{1mm}
\setlength{\tabcolsep}{6pt}
\renewcommand{\arraystretch}{1.05}
\begin{small}
\begin{tabular}{r cccc cc}
\toprule
Method & Am & Ca & D & W & AVG $\uparrow$ & STD $\downarrow$ \\ 
\midrule
FedAvg (CoOp) & 97.05 & 95.15 & 99.24 & 97.64 & 97.27 & 1.46 \\
+ \textbf{GGTPC} & 97.59 & 97.30 & 100.00& 100.00& \underline{\textbf{98.72}} & \underline{\textbf{1.28}} \\
\midrule
SCAFFOLD (CoOp) & 96.78 & 95.01 & 97.73 & 99.06 & 97.14 & 1.47 \\
+ \textbf{GGTPC} & 97.44 & 97.56 & 100.00& 100.00& \underline{\textbf{98.75}} & \underline{\textbf{1.25}} \\
\midrule
MOON (CoOp) & 97.45 & 94.20 & 98.48 & 98.58 & 97.18 & 1.77 \\
+ \textbf{GGTPC} & 97.74 & 96.92 & 100.00& 100.00& \underline{\textbf{98.67}} & 1.37 \\
\midrule
FedDyn (CoOp) & 97.18 & 94.74 & 99.24 & 98.58 & 97.44 & 1.72 \\
+ \textbf{GGTPC} & 97.59 & 97.18 & 100.00& 100.00& \underline{\textbf{98.69}} & \underline{\textbf{1.32}} \\
\midrule
FedOPT (CoOp) & 96.78 & 95.15 & 97.73 & 99.06 & 97.18 & 1.42 \\
+ \textbf{GGTPC} & 97.74 & 97.18 & 100.00& 100.00& \underline{\textbf{98.73}} & \underline{\textbf{1.29}} \\
\midrule
FedProto (CoOp) & 96.64 & 93.94 & 100.00& 98.58 & 97.29 & 2.27 \\
+ \textbf{GGTPC} & 97.44 & 97.43 & 100.00& 100.00& \underline{\textbf{98.72}} & \underline{\textbf{1.28}} \\
\midrule
FedNTD (CoOp) & 97.05 & 95.28 & 98.48 & 97.64 & 97.11 & \underline{\textbf{1.17}} \\
+ \textbf{GGTPC} & 97.74 & 96.92 & 100.00& 100.00& \underline{\textbf{98.67}} & 1.37 \\
\bottomrule
\end{tabular}
\end{small}
\label{tab:5}
\vspace{-3mm}
\end{table*}

\begin{table*}[t]
\centering
\caption{Performance evaluation on the PACS-LDS dataset. AP, Ct, P, and Sk denote Art Painting, Cartoon, Photo, and Sketch domains, respectively.}
\vspace{1mm}
\setlength{\tabcolsep}{6pt}
\renewcommand{\arraystretch}{1.05}
\begin{small}
\begin{tabular}{r cccc cc}
\toprule
Method & AP & Ct & P & Sk & AVG $\uparrow$ & STD $\downarrow$ \\ 
\midrule
FedAvg (CoOp) & 98.41 & 98.84 & 99.43 & 90.20 & 96.72 & 3.78 \\
+ \textbf{GGTPC} & 99.37 & 99.88 & 100.00& 96.33 & \underline{\textbf{98.90}} & \underline{\textbf{1.50}} \\
\midrule
SCAFFOLD (CoOp) & 98.49 & 99.02 & 99.76 & 91.42 & 97.17 & 3.35 \\
+ \textbf{GGTPC} & 99.44 & 99.82 & 99.91 & 96.80 & \underline{\textbf{98.99}} & \underline{\textbf{1.28}} \\
\midrule
MOON (CoOp) & 98.19 & 98.97 & 99.68 & 91.82 & 97.16 & 3.13 \\
+ \textbf{GGTPC} & 99.51 & 99.76 & 100.00& 96.91 & \underline{\textbf{99.05}} & \underline{\textbf{1.24}} \\
\midrule
FedDyn (CoOp) & 97.36 & 98.63 & 99.68 & 93.04 & 97.18 & 2.52 \\
+ \textbf{GGTPC} & 99.30 & 99.82 & 100.00& 96.44 & \underline{\textbf{98.89}} & \underline{\textbf{1.44}} \\
\midrule
FedOPT (CoOp) & 98.34 & 99.06 & 99.51 & 89.28 & 96.55 & 4.22 \\
+ \textbf{GGTPC} & 99.44 & 99.82 & 100.00& 96.77 & \underline{\textbf{99.01}} & \underline{\textbf{1.31}} \\
\midrule
FedProto (CoOp) & 98.11 & 98.89 & 99.76 & 91.32 & 97.02 & 3.34 \\
+ \textbf{GGTPC} & 99.44 & 99.82 & 100.00& 96.80 & \underline{\textbf{99.02}} & \underline{\textbf{1.29}} \\
\midrule
FedNTD (CoOp) & 97.73 & 98.93 & 99.84 & 89.99 & 96.62 & 3.90 \\
+ \textbf{GGTPC} & 99.03 & 99.88 & 100.00& 96.19 & \underline{\textbf{98.77}} & 1.54 \\
\bottomrule
\end{tabular}
\end{small}
\label{tab:6}
\vspace{-3mm}
\end{table*}

\textbf{Mixed Label and Domain Skew (LDS) Datasets:} To simulate more realistic and challenging scenarios, we construct two datasets exhibiting concurrent label and domain skew: Office-Caltech-LDS \cite{gong2012geodesic} and PACS-LDS \cite{li2017deeper}. For a dataset with $K$ domains and $C$ classes, we first assign $K$ clients to $K$ distinct domains and then generate a $K \times C$ coefficient matrix via a Dirichlet distribution $\text{Dir}(\beta)$\cite{fedavgm} to control the proportion of samples for each class on each client. This construction ensures that each client confronts both domain shift and local class imbalance.

\textbf{Implementation Details:} We employ CoOp as the backbone network for prompt learning, utilizing a pre-trained CLIP model based on the ViT-B/16 architecture. The prompt vector length is set to $L=16$. The optimizer employed is SGD, with a learning rate of $0.002$, weight decay of $1 \times 10^{-5}$, momentum of $0.9$, and a training batch size of $32$. Our method demonstrates high communication efficiency, requiring only $T=10$ communication rounds for single-domain tasks and $T=30$ rounds for multi-domain tasks.

\textbf{Evaluation Metrics:} We utilize Top-1 Accuracy as the primary metric to evaluate the generalization performance of the final global model. For multi-domain tasks, we also report the Standard Deviation (STD) of accuracy across different domains. A lower STD indicates superior cross-domain fairness and robustness of the model.

\textbf{Comparison Methods:} In single-domain (label skew) experiments, GGTPC is compared against representative federated prompt learning methods: PromptFL\cite{guo2023promptfl}, FedTPG\cite{qiu2024federated}, FedPR\cite{feng2023learning}, and FedCLIP\cite{fedclip} (plus FedVPT). In multi-domain experiments, we apply GGTPC as a plug-in to classic federated algorithms (FedAvg\cite{mcmahan2017communication}, SCAFFOLD\cite{SCAFFOLD}, MOON\cite{moon}, FedDyn\cite{feddyn}, FedOPT\cite{reddi2020adaptive}, FedProto\cite{fedproto}, and FedNTD\cite{FedNTD}).

\subsection{Results under Label Skew Scenarios}

\subsubsection{Performance Comparison with State-of-the-Art Methods}

To validate the fundamental effectiveness of the proposed GGTPC method, we first compared it with various mainstream Federated Prompt Learning methods on the CIFAR-100 and Tiny-ImageNet datasets. As shown in Table~\ref{tab:1}, the experimental results demonstrate that the proposed GGTPC method consistently outperforms all comparison methods under varying degrees of label skew ($\beta$ values ranging from 0.5 to 0.1). 

Taking the CIFAR-100 dataset as an example, when the data distribution is relatively balanced ($\beta=0.5$), GGTPC achieves an accuracy of $84.21\%$, which represents an improvement over the second-best performing method, FedVPT ($83.53\%$). As the degree of heterogeneity intensifies, the advantage of GGTPC becomes increasingly significant. Under the most challenging setting of $\beta=0.1$, the performance of the baseline method FedAvg (CoOp) drops to $75.92\%$, whereas GGTPC maintains an accuracy of $83.14\%$. This corresponds to a $7.22$ percentage point improvement over the baseline and a $2.15$ percentage point lead over the second-best method, FedVPT ($80.99\%$). A similar trend is observed on the Tiny-ImageNet dataset; at $\beta=0.1$, the performance of GGTPC ($79.26\%$) improves by $5.57\%$ compared to the baseline ($73.69\%$). These results compellingly prove that by introducing global geometric priors to calibrate local training, GGTPC effectively combats the distribution shift caused by label skew, thereby enabling the learning of more accurate and robust global prompts.

\subsubsection{Robustness Analysis under Extreme Skew Environments}

To further investigate the performance boundaries of the GGTPC method, we conducted experiments under a series of more severe label skew environments ($\beta$ values decreasing from 0.09 to 0.01). The results, presented in Table~\ref{tab:2}, reveal a critical phenomenon: as data heterogeneity increases drastically, the performance advantage of GGTPC over baseline methods is further amplified.

On the CIFAR-100 dataset, when $\beta$ decays from 0.05 to 0.01, the accuracy of the baseline FedAvg (CoOp) plummets from $72.99\%$ to $69.71\%$, suffering severe performance loss. In sharp contrast, GGTPC exhibits high stability, with accuracy declining only slightly from $79.63\%$ to $78.88\%$. This implies that under the extreme setting of $\beta=0.01$, GGTPC achieves a performance gain of up to $9.17$ percentage points relative to the baseline. This pattern is also replicated on CIFAR-10 and Tiny-ImageNet. 

This phenomenon profoundly validates the core motivation of this paper: when clients' local data views are extremely narrow and partial (i.e., extremely small $\beta$ values), the calculated local feature centers are far removed from the true global centers. In such instances, the global geometric prior knowledge provided by the server acts as a "God's eye view" calibrator, making its guidance crucial. This fully demonstrates the robust capabilities of the GGTPC method in handling extreme non-IID data distributions.

\subsection{Results under Domain Skew Scenarios}

To examine the compatibility of GGTPC as a universal module with different Federated Learning algorithms, we conducted a series of "plug-and-play" experiments on two domain-skewed datasets, Digits and Office-Home. In these experiments, GGTPC was integrated into seven classic Federated Learning algorithms, respectively. The results are presented in Table~\ref{tab:3} and Table~\ref{tab:4}.

The results in Table~\ref{tab:3} (Digits dataset) and Table~\ref{tab:4} (Office-Home dataset) consistently demonstrate that GGTPC yields stable performance gains for all tested Federated Learning algorithms. On the more complex Office-Home dataset, integrating GGTPC into FedAvg, SCAFFOLD, and MOON increased their average accuracy from $85.01\%$, $85.36\%$, and $85.40\%$ to $89.61\%$, $90.05\%$, and $89.49\%$, respectively, achieving significant improvements of $4.60$, $4.69$, and $4.09$ percentage points.

More noteworthy is the variation in the Standard Deviation (STD) of accuracy. On the Office-Home dataset, the STD for all baseline methods ranged between $6.0$ and $7.5$, whereas after integrating GGTPC, the STD generally decreased to the range of $5.1$ to $5.7$. A lower STD indicates that the model exhibits more balanced performance across different domains, avoiding overfitting to specific "dominant" domains, thereby enhancing the overall fairness and robustness of the model. This result indicates that by providing shared global geometric priors across domains, GGTPC helps individual single-domain clients "perceive" the data distribution morphology of other domains, effectively mitigating the knowledge gap caused by domain shift. These experiments fully demonstrate the powerful versatility and compatibility of GGTPC, establishing it as an effective enhancement module capable of empowering various Federated Learning frameworks.

\subsection{Results under Mixed Skew (LDS) Scenarios}

To simulate the complex conditions closest to real-world applications, we constructed a Mixed Skew (LDS) scenario exhibiting concurrent label and domain skew, and evaluated GGTPC on the Office-Caltech-LDS and PACS-LDS datasets. The results are shown in Table~\ref{tab:5} and Table~\ref{tab:6}.

Under such rigorous settings, GGTPC demonstrates immense potential for application. Taking the Office-Caltech-LDS dataset (Table~\ref{tab:5}) as an example, although the baseline FedAvg (CoOp) already achieved a high accuracy of $97.27\%$, leaving extremely limited room for improvement, the integration of GGTPC still boosted the accuracy to $98.72\%$, a gain of $1.45$ percentage points, while reducing the STD from $1.46$ to $1.28$. GGTPC delivered similar improvements across all tested algorithms, with average accuracies consistently exceeding $98.6\%$.

On the more challenging PACS-LDS dataset (Table~\ref{tab:6}), where baseline performance is more heavily impacted by substantial inter-domain discrepancies (e.g., Photo domain vs Sketch domain), the improvement effect of GGTPC is even more pronounced. For instance, when applied to FedAvg (CoOp), the average accuracy increased from $96.72\%$ to $98.90\%$, a gain of $2.18$ percentage points, while the STD was drastically compressed from $3.78$ to $1.50$. These results compellingly prove the success of our strategy combining geometric priors and class prototypes. It not only calibrates the distributional shape bias caused by class imbalance but also compensates for the lack of positional information caused by domain shift, thereby enabling the learning of high-quality global prompts even in the most complex heterogeneous environments.

\subsection{Visualization Analysis}

\begin{figure}[t]
    \centering
    \includegraphics[width=\linewidth,height=0.4\textheight,keepaspectratio]{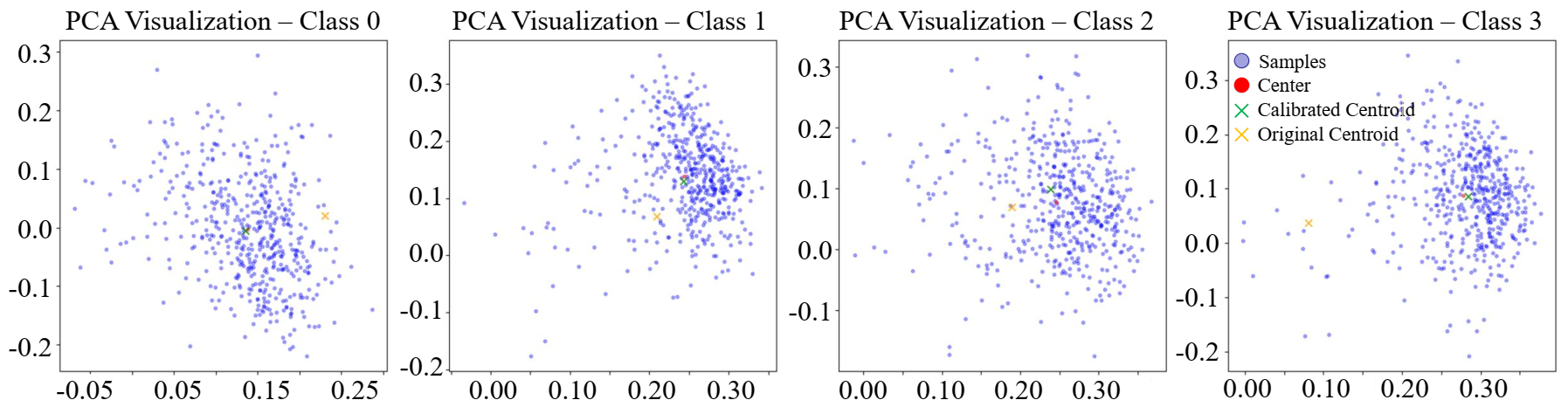}
    \caption{PCA visualization of global visual features and text prompt embeddings before/after GGTPC calibration across multiple classes.}
    \label{fig:vis}
\end{figure}

To provide qualitative and intuitive evidence for the core argument of this paper—that GGTPC can calibrate local training bias induced by data heterogeneity—this section performs a visualization analysis of text prompt embeddings and visual feature distributions. We employed Principal Component Analysis (PCA) to project the global visual feature embeddings of multiple classes from the test set, their global means, and the text prompt embeddings before and after GGTPC calibration into a shared two-dimensional space. The results are illustrated in Figure~\ref{fig:vis}.

The visualization results clearly reveal the effectiveness of our method. In the subplots for all classes, the blue scatter clouds constitute the true distribution morphology of the global visual features for that class, with their center (Red Point) being the theoretically optimal alignment target for the text prompt embedding. It can be observed that in all subplots, the text embeddings before calibration (Orange Cross) deviate significantly from their corresponding global visual feature centers (Red Point). In Class 0 and Class 3, this deviation is particularly substantial, with the Orange Cross even falling into sparse regions of the feature distribution. This phenomenon intuitively corroborates the core problem raised in our Introduction: in data-heterogeneous environments, baseline models trained solely on local data converge to a biased position and fail to accurately characterize the global data.

In sharp contrast, the calibrated text embeddings (Green Cross) align precisely with the centers of the global visual features across all classes. The Green Cross almost completely overlaps with the Red Point, sitting firmly in the densest region of the feature distribution. This indicates that by receiving global geometric priors and utilizing GPCL for local calibration, clients can effectively overcome the limited view of local data, successfully correcting the optimization direction of text prompts from biased local objectives to unbiased global objectives.

In summary, this visualization analysis provides strong evidence for the internal mechanism of GGTPC from the geometric perspective of feature space distribution, proving that it does not merely achieve improvements in classification accuracy but fundamentally resolves the embedding bias issue caused by data heterogeneity in Federated Prompt Learning.

\section{Conclusion}
\label{sec: conclusion}

This paper addresses the core challenge of local training bias induced by data heterogeneity in Federated Prompt Learning (FPL) by proposing a novel method named Geometry-Guided Text Prompt Calibration (GGTPC). Transcending the limitations of traditional server-side aggregation optimization or client-side regularization, our approach adopts a fundamentally new perspective, mitigating the distribution shift problem at its source by directly calibrating the client's local training process. The core mechanism of GGTPC involves aggregating local statistics on the server in a privacy-preserving manner to reconstruct the geometric shape of the global data distribution, which is then distributed to clients as prior knowledge. Subsequently, clients leverage the proposed Geometric Prior Calibration Layer (GPCL) to perform efficient end-to-end feature space calibration, thereby steering the optimization of text prompts toward unbiased global objectives.

Extensive experimental results substantiate the effectiveness and versatility of GGTPC. Across diverse and rigorous data heterogeneity scenarios—including label skew, domain skew, and mixed skew GGTPC not only exhibits superior standalone performance but also functions as a plug-and-play module that consistently bolsters the accuracy and fairness of various mainstream federated learning algorithms. Visualization analysis further provides intuitive geometric evidence that GGTPC successfully realigns biased text prompt embeddings with the true centers of global visual features.

In summary, this work offers a novel paradigm for addressing data heterogeneity in federated learning: shifting the focus from rectifying aggregation outcomes to calibrating the training source. By directly addressing distribution shifts at the feature level, GGTPC paves a new avenue for the development of more robust and efficient Federated Prompt Learning systems. Future research directions include exploring online update mechanisms for dynamic geometric priors to adapt to temporal distribution shifts, as well as extending this calibration philosophy to federated multi-modal learning or generative tasks.

\clearpage
\bibliographystyle{unsrt}
\bibliography{reference}

@String(NIPS = {Adv. Neural Inform. Process. Syst.})

@String(ICME = {Int. Conf. Multimedia and Expo})

@String(AAAI = {AAAI})

@inproceedings{mcmahan2017communication,
  title={Communication-efficient learning of deep networks from decentralized data},
  author={McMahan, Brendan and Moore, Eider and Ramage, Daniel and Hampson, Seth and y Arcas, Blaise Aguera},
  booktitle={Artificial intelligence and statistics},
  pages={1273--1282},
  year={2017},
  organization={PMLR}
}

@article{yang2019federated,
  title={Federated machine learning: Concept and applications},
  author={Yang, Qiang and Liu, Yang and Chen, Tianjian and Tong, Yongxin},
  journal={ACM Transactions on Intelligent Systems and Technology (TIST)},
  volume={10},
  number={2},
  pages={1--19},
  year={2019},
  publisher={ACM New York, NY, USA}
}

@article{kairouz2021advances,
  title={Advances and open problems in federated learning},
  author={Kairouz, Peter and McMahan, H Brendan and Avent, Brendan and Bellet, Aur{\'e}lien and Bennis, Mehdi and Bhagoji, Arjun Nitin and Bonawitz, Kallista and Charles, Zachary and Cormode, Graham and Cummings, Rachel and others},
  journal={Foundations and trends{\textregistered} in machine learning},
  volume={14},
  number={1--2},
  pages={1--210},
  year={2021},
  publisher={Now Publishers, Inc.}
}

@inproceedings{li2022federated,
  title={Federated learning on non-iid data silos: An experimental study},
  author={Li, Qinbin and Diao, Yiqun and Chen, Quan and He, Bingsheng},
  booktitle={2022 IEEE 38th international conference on data engineering (ICDE)},
  pages={965--978},
  year={2022},
  organization={IEEE}
}

@article{huang2024federated,
  title={Federated learning for generalization, robustness, fairness: A survey and benchmark},
  author={Huang, Wenke and Ye, Mang and Shi, Zekun and Wan, Guancheng and Li, He and Du, Bo and Yang, Qiang},
  journal={IEEE Transactions on Pattern Analysis and Machine Intelligence},
  year={2024},
  publisher={IEEE}
}

@article{zhou2022learning,
  title={Learning to prompt for vision-language models},
  author={Zhou, Kaiyang and Yang, Jingkang and Loy, Chen Change and Liu, Ziwei},
  journal={International Journal of Computer Vision},
  volume={130},
  number={9},
  pages={2337--2348},
  year={2022},
  publisher={Springer}
}

@inproceedings{zhou2022conditional,
  title={Conditional prompt learning for vision-language models},
  author={Zhou, Kaiyang and Yang, Jingkang and Loy, Chen Change and Liu, Ziwei},
  booktitle={Proceedings of the IEEE/CVF conference on computer vision and pattern recognition},
  pages={16816--16825},
  year={2022}
}

@article{zang2022unified,
  title={Unified vision and language prompt learning},
  author={Zang, Yuhang and Li, Wei and Zhou, Kaiyang and Huang, Chen and Loy, Chen Change},
  journal={arXiv preprint arXiv:2210.07225},
  year={2022}
}

@inproceedings{li2023efficient,
  title={Efficient multimodal fusion via interactive prompting},
  author={Li, Yaowei and Quan, Ruijie and Zhu, Linchao and Yang, Yi},
  booktitle={Proceedings of the IEEE/CVF conference on computer vision and pattern recognition},
  pages={2604--2613},
  year={2023}
}

@inproceedings{khattak2023maple,
  title={Maple: Multi-modal prompt learning},
  author={Khattak, Muhammad Uzair and Rasheed, Hanoona and Maaz, Muhammad and Khan, Salman and Khan, Fahad Shahbaz},
  booktitle={Proceedings of the IEEE/CVF conference on computer vision and pattern recognition},
  pages={19113--19122},
  year={2023}
}

@inproceedings{bulat2023lasp,
  title={Lasp: Text-to-text optimization for language-aware soft prompting of vision \& language models},
  author={Bulat, Adrian and Tzimiropoulos, Georgios},
  booktitle={Proceedings of the IEEE/CVF conference on computer vision and pattern recognition},
  pages={23232--23241},
  year={2023}
}

@inproceedings{yao2023visual,
  title={Visual-language prompt tuning with knowledge-guided context optimization},
  author={Yao, Hantao and Zhang, Rui and Xu, Changsheng},
  booktitle={Proceedings of the IEEE/CVF conference on computer vision and pattern recognition},
  pages={6757--6767},
  year={2023}
}

@inproceedings{zhang2024dept,
  title={Dept: Decoupled prompt tuning},
  author={Zhang, Ji and Wu, Shihan and Gao, Lianli and Shen, Heng Tao and Song, Jingkuan},
  booktitle={Proceedings of the IEEE/CVF Conference on Computer Vision and Pattern Recognition},
  pages={12924--12933},
  year={2024}
}

@article{guo2023promptfl,
  title={Promptfl: Let federated participants cooperatively learn prompts instead of models--federated learning in age of foundation model},
  author={Guo, Tao and Guo, Song and Wang, Junxiao and Tang, Xueyang and Xu, Wenchao},
  journal={IEEE Transactions on Mobile Computing},
  volume={23},
  number={5},
  pages={5179--5194},
  year={2023},
  publisher={IEEE}
}

@inproceedings{feng2023learning,
  title={Learning federated visual prompt in null space for mri reconstruction},
  author={Feng, Chun-Mei and Li, Bangjun and Xu, Xinxing and Liu, Yong and Fu, Huazhu and Zuo, Wangmeng},
  booktitle={Proceedings of the IEEE/CVF Conference on Computer Vision and Pattern Recognition},
  pages={8064--8073},
  year={2023}
}

@inproceedings{qiu2024federated,
  title={Federated text-driven prompt generation for vision-language models},
  author={Qiu, Chen and Li, Xingyu and Mummadi, Chaithanya Kumar and Ganesh, Madan Ravi and Li, Zhenzhen and Peng, Lu and Lin, Wan-Yi},
  booktitle={The Twelfth International Conference on Learning Representations},
  year={2024}
}

@article{li2020federated,
  title={Federated optimization in heterogeneous networks},
  author={Li, Tian and Sahu, Anit Kumar and Zaheer, Manzil and Sanjabi, Maziar and Talwalkar, Ameet and Smith, Virginia},
  journal={Proceedings of Machine learning and systems},
  volume={2},
  pages={429--450},
  year={2020}
}

@article{SCAFFOLD,
  title={Scaffold: Stochastic controlled averaging for on-device federated learning},
  author={Karimireddy, Sai Praneeth and Kale, Satyen and Mohri, Mehryar and Reddi, Sashank J and Stich, Sebastian U and Suresh, Ananda Theertha},
  journal={arXiv preprint arXiv:1910.06378},
  volume={2},
  number={6},
  year={2019}
}

@article{reddi2020adaptive,
  title={Adaptive federated optimization},
  author={Reddi, Sashank and Charles, Zachary and Zaheer, Manzil and Garrett, Zachary and Rush, Keith and Kone{\v{c}}n{\`y}, Jakub and Kumar, Sanjiv and McMahan, H Brendan},
  journal={arXiv preprint arXiv:2003.00295},
  year={2020}
}

@article{krizhevsky2009learning,
  title={Learning multiple layers of features from tiny images},
  author={Krizhevsky, Alex and Hinton, Geoffrey and others},
  journal={Technical Report},
  year={2009},
  publisher={Toronto, ON, Canada}
}

@inproceedings{deng2009imagenet,
  title={Imagenet: A large-scale hierarchical image database},
  author={Deng, Jia and Dong, Wei and Socher, Richard and Li, Li-Jia and Li, Kai and Fei-Fei, Li},
  booktitle={2009 IEEE conference on computer vision and pattern recognition},
  pages={248--255},
  year={2009},
  organization={Ieee}
}

@article{hull1994database,
  title={A database for handwritten text recognition research},
  author={Hull, Jonathan J.},
  journal={IEEE Transactions on pattern analysis and machine intelligence},
  volume={16},
  number={5},
  pages={550--554},
  year={1994},
  publisher={IEEE}
}

@article{lecun1998gradient,
  title={Gradient-based learning applied to document recognition},
  author={LeCun, Yann and Bottou, L{\'e}on and Bengio, Yoshua and Haffner, Patrick},
  journal={Proceedings of the IEEE},
  volume={86},
  number={11},
  pages={2278--2324},
  year={1998},
  publisher={Ieee}
}

@inproceedings{netzer2011reading,
  title={Reading digits in natural images with unsupervised feature learning},
  author={Netzer, Yuval and Wang, Tao and Coates, Adam and Bissacco, Alessandro and Wu, Baolin and Ng, Andrew Y and others},
  booktitle={NIPS workshop on deep learning and unsupervised feature learning},
  volume={2011},
  pages={4},
  year={2011},
  organization={Granada}
}

@inproceedings{li2017deeper,
  title={Deeper, broader and artier domain generalization},
  author={Li, Da and Yang, Yongxin and Song, Yi-Zhe and Hospedales, Timothy M},
  booktitle={Proceedings of the IEEE international conference on computer vision},
  pages={5542--5550},
  year={2017}
}

@inproceedings{ganin2015unsupervised,
  title={Unsupervised domain adaptation by backpropagation},
  author={Ganin, Yaroslav and Lempitsky, Victor},
  booktitle={International conference on machine learning},
  pages={1180--1189},
  year={2015},
  organization={PMLR}
}

@inproceedings{gong2012geodesic,
  title={Geodesic flow kernel for unsupervised domain adaptation},
  author={Gong, Boqing and Shi, Yuan and Sha, Fei and Grauman, Kristen},
  booktitle={2012 IEEE conference on computer vision and pattern recognition},
  pages={2066--2073},
  year={2012},
  organization={IEEE}
}

@inproceedings{venkateswara2017deep,
  title={Deep hashing network for unsupervised domain adaptation},
  author={Venkateswara, Hemanth and Eusebio, Jose and Chakraborty, Shayok and Panchanathan, Sethuraman},
  booktitle={Proceedings of the IEEE conference on computer vision and pattern recognition},
  pages={5018--5027},
  year={2017}
}

@inproceedings{moon,
  title={Model-contrastive federated learning},
  author={Li, Qinbin and He, Bingsheng and Song, Dawn},
  booktitle={Proceedings of the IEEE/CVF conference on computer vision and pattern recognition},
  pages={10713--10722},
  year={2021}
}

@article{feddyn,
  title={Federated learning based on dynamic regularization},
  author={Acar, Durmus Alp Emre and Zhao, Yue and Navarro, Ramon Matas and Mattina, Matthew and Whatmough, Paul N and Saligrama, Venkatesh},
  journal={arXiv preprint arXiv:2111.04263},
  year={2021}
}

@InProceedings{CLIP,
  title = 	 {Learning Transferable Visual Models From Natural Language Supervision},
  author =       {Radford, Alec and Kim, Jong Wook and Hallacy, Chris and Ramesh, Aditya and Goh, Gabriel and Agarwal, Sandhini and Sastry, Girish and Askell, Amanda and Mishkin, Pamela and Clark, Jack and Krueger, Gretchen and Sutskever, Ilya},
  booktitle = 	 {Proceedings of the 38th International Conference on Machine Learning},
  pages = 	 {8748--8763},
  year = 	 {2021},
  editor = 	 {Meila, Marina and Zhang, Tong},
  volume = 	 {139},
  series = 	 {Proceedings of Machine Learning Research},
  month = 	 {18--24 Jul},
  publisher =    {PMLR},
  url = 	 {https://proceedings.mlr.press/v139/radford21a.html}
}

@article{ye2023heterogeneous,
  title={Heterogeneous federated learning: State-of-the-art and research challenges},
  author={Ye, Mang and Fang, Xiuwen and Du, Bo and Yuen, Pong C and Tao, Dacheng},
  journal={ACM Computing Surveys},
  volume={56},
  number={3},
  pages={1--44},
  year={2023},
  publisher={ACM New York, NY, USA}
}

@article{zhao2018federated,
  title={Federated learning with non-iid data},
  author={Zhao, Yue and Li, Meng and Lai, Liangzhen and Suda, Naveen and Civin, Damon and Chandra, Vikas},
  journal={arXiv preprint arXiv:1806.00582},
  year={2018}
}

@inproceedings{gao2022feddc,
  title={Feddc: Federated learning with non-iid data via local drift decoupling and correction},
  author={Gao, Liang and Fu, Huazhu and Li, Li and Chen, Yingwen and Xu, Ming and Xu, Cheng-Zhong},
  booktitle={Proceedings of the IEEE/CVF conference on computer vision and pattern recognition},
  pages={10112--10121},
  year={2022}
}

@article{yoon2021fedmix,
  title={Fedmix: Approximation of mixup under mean augmented federated learning},
  author={Yoon, Tehrim and Shin, Sumin and Hwang, Sung Ju and Yang, Eunho},
  journal={arXiv preprint arXiv:2107.00233},
  year={2021}
}

@inproceedings{fedrs,
  title={Fedrs: Federated learning with restricted softmax for label distribution non-iid data},
  author={Li, Xin-Chun and Zhan, De-Chuan},
  booktitle={Proceedings of the 27th ACM SIGKDD conference on knowledge discovery \& data mining},
  pages={995--1005},
  year={2021}
}

@inproceedings{fedlc,
  title={Federated learning with label distribution skew via logits calibration},
  author={Zhang, Jie and Li, Zhiqi and Li, Bo and Xu, Jianghe and Wu, Shuang and Ding, Shouhong and Wu, Chao},
  booktitle={International Conference on Machine Learning},
  pages={26311--26329},
  year={2022},
  organization={PMLR}
}

@inproceedings{fedproto,
  title={Fedproto: Federated prototype learning across heterogeneous clients},
  author={Tan, Yue and Long, Guodong and Liu, Lu and Zhou, Tianyi and Lu, Qinghua and Jiang, Jing and Zhang, Chengqi},
  booktitle={Proceedings of the AAAI Conference on Artificial Intelligence},
  volume={36},
  number={8},
  pages={8432--8440},
  year={2022}
}

@article{fednova,
  title={Tackling the objective inconsistency problem in heterogeneous federated optimization},
  author={Wang, Jianyu and Liu, Qinghua and Liang, Hao and Joshi, Gauri and Poor, H Vincent},
  journal={Advances in neural information processing systems},
  volume={33},
  pages={7611--7623},
  year={2020}
}

@article{fedavgm,
  title={Measuring the effects of non-identical data distribution for federated visual classification},
  author={Hsu, Tzu-Ming Harry and Qi, Hang and Brown, Matthew},
  journal={arXiv preprint arXiv:1909.06335},
  year={2019}
}

@article{fedclip,
  title={Fedclip: Fast generalization and personalization for clip in federated learning},
  author={Lu, Wang and Hu, Xixu and Wang, Jindong and Xie, Xing},
  journal={arXiv preprint arXiv:2302.13485},
  year={2023}
}

@article{li2020federated2,
  title={Federated learning: Challenges, methods, and future directions},
  author={Li, Tian and Sahu, Anit Kumar and Talwalkar, Ameet and Smith, Virginia},
  journal={IEEE signal processing magazine},
  volume={37},
  number={3},
  pages={50--60},
  year={2020},
  publisher={IEEE}
}

@article{li2021survey,
  title={A survey on federated learning systems: Vision, hype and reality for data privacy and protection},
  author={Li, Qinbin and Wen, Zeyi and Wu, Zhaomin and Hu, Sixu and Wang, Naibo and Li, Yuan and Liu, Xu and He, Bingsheng},
  journal={IEEE Transactions on Knowledge and Data Engineering},
  volume={35},
  number={4},
  pages={3347--3366},
  year={2021},
  publisher={IEEE}
}

@article{yin2021comprehensive,
  title={A comprehensive survey of privacy-preserving federated learning: A taxonomy, review, and future directions},
  author={Yin, Xuefei and Zhu, Yanming and Hu, Jiankun},
  journal={ACM Computing Surveys (CSUR)},
  volume={54},
  number={6},
  pages={1--36},
  year={2021},
  publisher={ACM New York, NY, USA}
}

@article{bonawitz2019towards,
  title={Towards federated learning at scale: System design},
  author={Bonawitz, Keith and Eichner, Hubert and Grieskamp, Wolfgang and Huba, Dzmitry and Ingerman, Alex and Ivanov, Vladimir and Kiddon, Chloe and Kone{\v{c}}n{\`y}, Jakub and Mazzocchi, Stefano and McMahan, Brendan and others},
  journal={Proceedings of machine learning and systems},
  volume={1},
  pages={374--388},
  year={2019}
}

@inproceedings{BLIP,
  title={Blip: Bootstrapping language-image pre-training for unified vision-language understanding and generation},
  author={Li, Junnan and Li, Dongxu and Xiong, Caiming and Hoi, Steven},
  booktitle={International conference on machine learning},
  pages={12888--12900},
  year={2022},
  organization={PMLR}
}

@inproceedings{BLIP2,
  title={Blip-2: Bootstrapping language-image pre-training with frozen image encoders and large language models},
  author={Li, Junnan and Li, Dongxu and Savarese, Silvio and Hoi, Steven},
  booktitle={International conference on machine learning},
  pages={19730--19742},
  year={2023},
  organization={PMLR}
}

@inproceedings{hu2024fedmut,
  title={FedMut: Generalized federated learning via stochastic mutation},
  author={Hu, Ming and Cao, Yue and Li, Anran and Li, Zhiming and Liu, Chengwei and Li, Tianlin and Chen, Mingsong and Liu, Yang},
  booktitle={Proceedings of the AAAI conference on artificial intelligence},
  volume={38},
  number={11},
  pages={12528--12537},
  year={2024}
}

@inproceedings{hu2024fedcross,
  title={FedCross: Towards accurate federated learning via multi-model cross-aggregation},
  author={Hu, Ming and Zhou, Peiheng and Yue, Zhihao and Ling, Zhiwei and Huang, Yihao and Li, Anran and Liu, Yang and Lian, Xiang and Chen, Mingsong},
  booktitle={2024 IEEE 40th International Conference on Data Engineering (ICDE)},
  pages={2137--2150},
  year={2024},
  organization={IEEE}
}

@inproceedings{hu2024aggregation,
  title={Is aggregation the only choice? federated learning via layer-wise model recombination},
  author={Hu, Ming and Yue, Zhihao and Xie, Xiaofei and Chen, Cheng and Huang, Yihao and Wei, Xian and Lian, Xiang and Liu, Yang and Chen, Mingsong},
  booktitle={Proceedings of the 30th ACM SIGKDD Conference on Knowledge Discovery and Data Mining},
  pages={1096--1107},
  year={2024}
}

@inproceedings{qi2023cross,
  title={Cross-silo prototypical calibration for federated learning with non-iid data},
  author={Qi, Zhuang and Meng, Lei and Chen, Zitan and Hu, Han and Lin, Hui and Meng, Xiangxu},
  booktitle={Proceedings of the 31st ACM international conference on multimedia},
  pages={3099--3107},
  year={2023}
}

@inproceedings{qi2025cross,
  title={Cross-silo feature space alignment for federated learning on clients with imbalanced data},
  author={Qi, Zhuang and Meng, Lei and Li, Zhaochuan and Hu, Han and Meng, Xiangxu},
  booktitle={Proceedings of the AAAI Conference on Artificial Intelligence},
  volume={39},
  number={19},
  pages={19986--19994},
  year={2025}
}

@article{meng2024improving,
  title={Improving global generalization and local personalization for federated learning},
  author={Meng, Lei and Qi, Zhuang and Wu, Lei and Du, Xiaoyu and Li, Zhaochuan and Cui, Lizhen and Meng, Xiangxu},
  journal={IEEE Transactions on Neural Networks and Learning Systems},
  year={2024},
  publisher={IEEE}
}

@inproceedings{liu2023cross,
  title={Cross-training with prototypical distillation for improving the generalization of federated learning},
  author={Liu, Tianhan and Qi, Zhuang and Chen, Zitan and Meng, Xiangxu and Meng, Lei},
  booktitle={2023 IEEE International Conference on Multimedia and Expo (ICME)},
  pages={648--653},
  year={2023},
  organization={IEEE}
}

@inproceedings{qi2022clustering,
  title={Clustering-based curriculum construction for sample-balanced federated learning},
  author={Qi, Zhuang and Wang, Yuqing and Chen, Zitan and Wang, Ran and Meng, Xiangxu and Meng, Lei},
  booktitle={CAAI international conference on artificial intelligence},
  pages={155--166},
  year={2022},
  organization={Springer}
}

@article{zhao2024data,
  title={Data-free knowledge distillation via generator-free data generation for Non-IID federated learning},
  author={Zhao, Siran and Liao, Tianchi and Fu, Lele and Chen, Chuan and Bian, Jing and Zheng, Zibin},
  journal={Neural Networks},
  volume={179},
  pages={106627},
  year={2024},
  publisher={Elsevier}
}

@article{mai2024fgtl,
  title={FGTL: Federated graph transfer learning for node classification},
  author={Mai, Chengyuan and Liao, Tianchi and Chen, Chuan and Zheng, Zibin},
  journal={ACM Transactions on Knowledge Discovery from Data},
  volume={19},
  number={1},
  pages={1--20},
  year={2024},
  publisher={ACM New York, NY}
}

@inproceedings{wang2023dafkd,
  title={Dafkd: Domain-aware federated knowledge distillation},
  author={Wang, Haozhao and Li, Yichen and Xu, Wenchao and Li, Ruixuan and Zhan, Yufeng and Zeng, Zhigang},
  booktitle={Proceedings of the IEEE/CVF conference on computer vision and pattern recognition},
  pages={20412--20421},
  year={2023}
}

@inproceedings{wang2024fedcda,
  title={Fedcda: Federated learning with cross-rounds divergence-aware aggregation},
  author={Wang, Haozhao and Xu, Haoran and Li, Yichen and Xu, Yuan and Li, Ruixuan and Zhang, Tianwei},
  booktitle={The Twelfth International Conference on Learning Representations},
  year={2024}
}

@inproceedings{ma2025geometric,
  title={Geometric knowledge-guided localized global distribution alignment for federated learning},
  author={Ma, Yanbiao and Dai, Wei and Huang, Wenke and Chen, Jiayi},
  booktitle={Proceedings of the Computer Vision and Pattern Recognition Conference},
  pages={20958--20968},
  year={2025}
}

@article{qi2025oncology,
  title={Federated Deconfounding and Debiasing Learning for Out-of-Distribution Generalization},
  author={Qi, Zhuang and Zhou, Sijin and Meng, Lei and Hu, Han and Yu, Han and Meng, Xiangxu},
  journal={arXiv preprint arXiv:2505.04979},
  year={2025}
}

@article{qi2025global,
  title={Global Intervention and Distillation for Federated Out-of-Distribution Generalization},
  author={Qi, Zhuang and Zhang, Runhui and Meng, Lei and Wu, Wei and Zhang, Yachong and Meng, Xiangxu},
  journal={arXiv preprint arXiv:2504.00850},
  year={2025}
}

@inproceedings{qi2024attentive,
  title={Attentive modeling and distillation for out-of-distribution generalization of federated learning},
  author={Qi, Zhuang and He, Weihao and Meng, Xiangxu and Meng, Lei},
  booktitle={2024 IEEE International Conference on Multimedia and Expo (ICME)},
  pages={1--6},
  year={2024},
  organization={IEEE}
}

@article{zhang2025federated,
  title={Federated out-of-distribution generalization: A causal augmentation view},
  author={Zhang, Runhui and Zhou, Sijin and Qi, Zhuang},
  journal={arXiv preprint arXiv:2504.19882},
  year={2025}
}

@inproceedings{fu2025beyond,
  title={Beyond federated prototype learning: Learnable semantic anchors with hyperspherical contrast for domain-skewed data},
  author={Fu, Lele and Huang, Sheng and Lai, Yanyi and Liao, Tianchi and Zhang, Chuanfu and Chen, Chuan},
  booktitle={Proceedings of the AAAI Conference on Artificial Intelligence},
  volume={39},
  number={16},
  pages={16648--16656},
  year={2025}
}

@article{fu2025federated,
  title={Federated domain-independent prototype learning with alignments of representation and parameter spaces for feature shift},
  author={Fu, Lele and Huang, Sheng and Lai, Yanyi and Zhang, Chuanfu and Dai, Hong-Ning and Zheng, Zibin and Chen, Chuan},
  journal={IEEE Transactions on Mobile Computing},
  year={2025},
  publisher={IEEE}
}

@inproceedings{liao2025federated,
  title={Federated domain generalization with decision insight matrix},
  author={Liao, Tianchi and Xie, Binghui and Lele Fu, Sheng Huang and Deng, Bowen and Chen, Chuan and Zheng, Zibin},
  booktitle={Proceedings of the Thirty-Fourth International Joint Conference on Artificial Intelligence},
  pages={5689--5697},
  year={2025}
}

@article{chen2025advances,
  title={Advances in robust federated learning: A survey with heterogeneity considerations},
  author={Chen, Chuan and Liao, Tianchi and Deng, Xiaojun and Wu, Zihou and Huang, Sheng and Zheng, Zibin},
  journal={IEEE Transactions on Big Data},
  year={2025},
  publisher={IEEE}
}

@article{li2025unleashing,
  title={Unleashing the power of continual learning on non-centralized devices: A survey},
  author={Li, Yichen and Wang, Haozhao and Xu, Wenchao and Xiao, Tianzhe and Liu, Hong and Tu, Minzhu and Wang, Yuying and Yang, Xin and Zhang, Rui and Yu, Shui and others},
  journal={IEEE Communications Surveys \& Tutorials},
  year={2025},
  publisher={IEEE}
}

@article{li2025systematic,
  title={A systematic survey on federated sequential recommendation},
  author={Li, Yichen and Qin, Qiyu and Zhu, Gaoyang and Xu, Wenchao and Wang, Haozhao and Li, Yuhua and Zhang, Rui and Li, Ruixuan},
  journal={arXiv preprint arXiv:2504.05313},
  year={2025}
}

@article{qi2025science,
  title={Federated learning in oncology: bridging artificial intelligence innovation and privacy protection},
  author={Qi, Zhuang and Qi, Xin and Xu, Tao and Dang, Chengrun and Meng, Lei and Yu, Han},
  journal={Authorea Preprints},
  year={2025},
  publisher={Authorea}
}

@article{qi2025federated,
  title={Federated Learning for Science: A Survey on the Path to a Trustworthy Collaboration Ecosystem},
  author={Qi, Xin and Li, Meixuan and Zhou, Sijin and Feng, Wei and Qi, Zhuang},
  year={2025}
}

@inproceedings{fedvpt,
title={Visual prompt tuning},
author={Jia, Menglin and Tang, Luming and Chen, Bor-Chun and Cardie, Claire and Belongie, Serge and Hariharan, Bharath and Lim, Ser-Nam},
booktitle={European conference on computer vision},
pages={709--727},
year={2022},
organization={Springer}
}

@inproceedings{FedNTD,
title = "Preservation of the Global Knowledge by Not-True Distillation in Federated Learning",
author = "Gihun Lee and Minchan Jeong and Yongjin Shin and Sangmin Bae and Yun, \{Se Young\}",
note = "Publisher Copyright: {\textcopyright} 2022 Neural information processing systems foundation. All rights reserved.; 36th Conference on Neural Information Processing Systems, NeurIPS 2022 ; Conference date: 28-11-2022 Through 09-12-2022",
year = "2022",
language = "English",
series = "Advances in Neural Information Processing Systems",
publisher = "Neural information processing systems foundation",
editor = "S. Koyejo and S. Mohamed and A. Agarwal and D. Belgrave and K. Cho and A. Oh",
booktitle = "Advances in Neural Information Processing Systems 35 - 36th Conference on Neural Information Processing Systems, NeurIPS 2022",
}

\end{document}